%% file: ELM_GP_jrnl_2023.tex
\documentclass[pdflatex,sn-mathphys,Numbered]{sn-jnl}
\usepackage{setspace}
\usepackage{graphicx}%
\usepackage{multirow}%
\usepackage{amsmath,amssymb,amsfonts}%
\usepackage{amsthm}%
\usepackage{mathrsfs}%
\usepackage[title]{appendix}%
\usepackage{xcolor}%
\usepackage{textcomp}%
\usepackage{manyfoot}%
\usepackage{booktabs}%
\usepackage[linesnumbered,ruled,noend]{algorithm2e}

\usepackage{listings}%
\definecolor{mygray}{RGB}{240,240,240}
\usepackage{rotating}
\usepackage[many]{tcolorbox}
\usepackage{varwidth}
\usepackage{environ}

\usepackage{xspace}
\usepackage{paralist}
\usepackage{url}
\usepackage{subcaption}

\usepackage{colortbl}

\usepackage{tikz}
\usetikzlibrary{shapes.geometric, arrows}
\tikzstyle{arrow} = [thick,->,>=stealth]
\tikzstyle{operator} = [rectangle, rounded corners, minimum width=0.5cm, minimum height=0.5cm,text centered, text width=1.8cm, draw=black, fill=gray!40]
\tikzstyle{data} = [text centered]
\tikzstyle{LLM} = [rectangle, rounded corners, minimum width=0.5cm, minimum height=0.5cm,text centered, text width=1.1cm, draw=black, fill=gray!30]
\tikzstyle{prompt} = [rectangle, rounded corners, minimum width=0.5cm, minimum height=0.5cm,text centered, text width=1.1cm, draw=black, fill=gray!20]
\tikzstyle{truth} = [text centered]
\tikzstyle{answer} = [text centered, font={\small\bfseries}]
\tikzstyle{nL} = [anchor=north, yshift=-0.35cm]
\tikzstyle{eL} = [anchor=east]
\tikzstyle{nodeL} = [anchor=north]
\tikzset{
  font={\fontsize{7pt}{12}\selectfont}}

\newlength{\bubblesep}
\newlength{\bubblewidth}
\AtBeginDocument{\setlength{\bubblewidth}{.85\textwidth}}
\definecolor{bubblegreen}{RGB}{103,184,104}
\definecolor{bubblegray}{RGB}{241,240,240}

\newcommand{\bubble}[4]{%
  \tcbox[
    on line,
    arc=4.5mm,
    colback=#1,
    colframe=#1,
    #2,
  ]{\color{#3}\begin{varwidth}{\bubblewidth}#4\end{varwidth}}%
}

\ExplSyntaxOn
\seq_new:N \l__ooker_bubbles_seq
\tl_new:N \l__ooker_bubbles_first_tl
\tl_new:N \l__ooker_bubbles_last_tl

\NewEnviron{rightbubbles}
 {
  \begin{flushright}
  \ttfamily
  \seq_set_split:NnV \l__ooker_bubbles_seq { \par } \BODY
  \int_compare:nTF { \seq_count:N \l__ooker_bubbles_seq < 2 }
   {
    \bubble{bubblegreen}{rounded~corners}{white}{\BODY}\par
   }
   {
    \seq_pop_left:NN \l__ooker_bubbles_seq \l__ooker_bubbles_first_tl
    \seq_pop_right:NN \l__ooker_bubbles_seq \l__ooker_bubbles_last_tl
    \bubble{bubblegreen}{sharp~corners=southeast}{white}{\l__ooker_bubbles_first_tl}
    \par\nointerlineskip
    \addvspace{\bubblesep}
    \seq_map_inline:Nn \l__ooker_bubbles_seq
     {
      \bubble{bubblegreen}{sharp~corners=east}{white}{##1}
      \par\nointerlineskip
      \addvspace{\bubblesep}
     }
    \bubble{bubblegreen}{sharp~corners=northeast}{white}{\l__ooker_bubbles_last_tl}
    \par
   }
   \end{flushright}
 }
 \NewEnviron{leftbubbles}
 {
  \begin{flushleft}
  \ttfamily
  \seq_set_split:NnV \l__ooker_bubbles_seq { \par } \BODY
  \int_compare:nTF { \seq_count:N \l__ooker_bubbles_seq < 2 }
   {
    \bubble{bubblegray}{rounded~corners}{black}{\BODY}\par
   }
   {
    \seq_pop_left:NN \l__ooker_bubbles_seq \l__ooker_bubbles_first_tl
    \seq_pop_right:NN \l__ooker_bubbles_seq \l__ooker_bubbles_last_tl
    \bubble{bubblegray}{sharp~corners=southwest}{black}{\l__ooker_bubbles_first_tl}
    \par\nointerlineskip
    \addvspace{\bubblesep}
    \seq_map_inline:Nn \l__ooker_bubbles_seq
     {
      \bubble{bubblegray}{sharp~corners=west}{black}{##1}
      \par\nointerlineskip
      \addvspace{\bubblesep}
     }
    \bubble{bubblegray}{sharp~corners=northwest}{black}{\l__ooker_bubbles_last_tl}\par
   }
  \end{flushleft}
 }
 \ExplSyntaxOff

\newcommand{\ALFAECLLM}{\texttt{Tutorial-LLM\_GP}\xspace}

\newcommand{\LLM}{\textnormal{LLM}}
\newcommand{\GP}{Tutorial GP\xspace}

\newcommand{\GPLLM}{LLM\_GP\xspace}  

\newcommand{\GPSomeLLM}{LLM\_GP\_Mu\_XO\xspace}

\newcommand{\prims}{F\cup T}

\begin{document}

 \title{Evolving Code with A Large Language Model}

\author*[1]{\fnm{Erik} \sur{Hemberg}}\email{hembergerik@csail.mit.edu}
\author[1]{\fnm{Stephen} \sur{Moskal}}\email{smoskal@mit.edu}
\author[1]{\fnm{Una-May} \sur{O'Reilly}}\email{unamay@csail.mit.edu}
\affil*[1]{\orgdiv{EECS}, \orgname{MIT CSAIL}, \orgaddress{\street{32 Vassar St}, \city{Cambridge}, \postcode{02139}, \state{MA},  \country{USA}}}

\abstract{
Algorithms that use Large Language Models (LLMs) to evolve code  arrived on the Genetic Programming (GP) scene very recently.  
We present \GPLLM, a formalized LLM-based evolutionary algorithm designed to evolve code. 
Like GP, it uses evolutionary operators, but its designs and implementations of those operators radically differ from GP's because they enlist an LLM, using prompting and the LLM's pre-trained pattern matching and sequence completion capability. 
We also present a demonstration-level variant of \GPLLM and share its code.
By addressing algorithms that range from the formal to hands-on, we cover design and LLM-usage considerations as well as the scientific challenges that arise when using an LLM for genetic programming. 

}


\maketitle

\input{introduction}

\input{background}

\input{methods}

\input{experiments}

\input{discussion}

\input{conclusions}

\section*{Statements and Declarations}

The authors declare no competing interests.

The authors acknowledge funding for this work under US Government
Contract \#FA8075-18-D-0008.

Conceptualization: Erik Hemberg and Una-May O'Reilly; Methodology:
Erik Hemberg, Stephen Moskal and Una-May O'Reilly; Formal analysis and
investigation: Erik Hemberg; Writing - original draft preparation:
Erik Hemberg and Una-May O'Reilly; Writing - review and editing: Erik
Hemberg, Stephen Moskal and Una-May O'Reilly; Funding acquisition:
Una-May O'Reilly; Resources: Una-May O'Reilly; Supervision: Una-May
O'Reilly.

\input{appendix}

\bibliography{bibliography,llm_references}

\end{document}

%% file: introduction.tex
\section{Introduction}\label{sec:intro}


Large language models (LLMs), along with other Foundational Models, have disrupted conventional expectations of Artificial Intelligence systems. 
An LLM, with a chatbot or Natural Language API,  typically works in the input-output space of natural language, i.e. unstructured text. In a question-answer style, it processes natural language prompts and responds in natural language. 
Technically speaking, it is  a pre-trained transformer model\footnote{A type of deep neural network, see \cite{vaswani2017}.} which has distilled statistical patterns from a massive training set within its massive quantity of  numerical parameters and artificial neural architecture.  Pre-training is a process  which back-propagates errors arising from predictions that complete text sequences which come from massive training data. In many cases the model is then further fine-tuned on specifically selected data. 
Finally, a process called Reinforcement Learning with Human Feedback~\cite{griffith2013policy} is run to set up prompt-response (or question-answer) capability.  

In comparison, Evolutionary Algorithms (EA), inspired by Neo-Darwinian evolution, operate on a population of candidate solutions. Generally, an EA uses operators. It has one operator that initializes a population of candidate solutions, two others that iteratively test each solution and calculate its fitness, one that selects parents by referencing solution fitness, operators that vary parental copies randomly, and one to compose a new population that replaces the old. 
A basic EA is set up with its operators. Before execution of a run, it  is provided with a solution representation and a fitness function.  GP is an evolutionary algorithm, one that evolves code. It follows the general algorithmic structure of an EA while it has GP-specific versions of evolutionary operators such as crossover. 

Surprisingly, LLMs are able to generate code to solve many software engineering tasks and even program synthesis. And, both GP and LLMs are  \textit{able to evolve code}, i.e. perform program synthesis with an evolutionary-inspired method.
The LLM-basis of this claim is backed up by recent research with  noteworthy results. In very original work~\cite{bradley2023,chen2023evoprompting,liventsev2023fully}, open challenges in GP~\cite{o2010open,o2020automatic} are being addressed with
LLMs integrated with some sort of Evolutionary Algorithm. 
For example, an LLM has been enlisted to perform the function of different operators in the OpenELM Library~\cite{bradley2023}, 
an LLM has been used for code-level neural architecture search~\cite{chen2023evoprompting}, 
and, largely without using a typical GP algorithm, though by using  \textit{similar} evolutionary mechanisms, a large number of GP benchmarks for Automatic Programming   have been solved with an LLM-supported procedure~\cite{liventsev2023fully}.

\begin{figure}[h]
     \centering
     \includegraphics[width=\textwidth, bb=0 0 1000 1000]{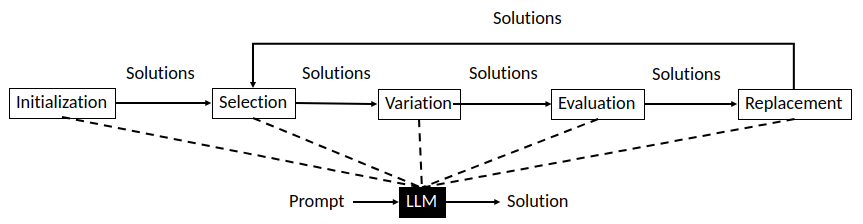}
     \caption{Overview of GP process with LLM operators. Codes are the population. Prompts are different for each LLM operator.}
     \label{fig:overview_GP_LLM}
\end{figure}

One objective of this paper is describe how an algorithm, with the general algorithmic structure of an EA and evolutionary operators, can use an LLM to evolve code, see Figure~\ref{fig:overview_GP_LLM}.  We describe how the operators are designed to formulate LLM prompts, task the LLM via the prompts, and process LLM responses, while code is represented as a sequence of text in code syntax.  The prompts ``task'' the LLM to fulfill the purpose of the operator: to initialize candidate solutions, select parents based on their performance, to vary one by mutation or more than one solution by recombination, etc.  The description is accompanied by operator and prompt design information and it presents preparatory run steps that are LLM-usage specific.
Another objective is to provide an implementation and demonstration of a simple \GPLLM variant. We hope to demystify the approach and provide a hands-on starting point for exploration.

We start, in Section~\ref{sec:background}, with background on LLMs and code models.  In Section~\ref{sec:spectrum} we introduce \GPLLM  and the design of its LLM-based operators, prompt-functions and LLM-oriented preparatory steps. This helps us, in Section~\ref{sec:related-work}, to  describe and contrast  the relatively  small body of current work integrating GP and LLMs. Moving to a hands-on perspective, in Section~\ref{sec:experiments} we provide a simple-to-understand \GPLLM variant within a open-source software package named \ALFAECLLM. Modules of the package contain both the algorithm's prompt functions and its prompts. These are a very sensitive part of using LLMs and frequently draw a lot of curiosity. We use this variant to demonstrate  time and financial costs, usage statistics, and errors related to LLM usage.  We present a discussion in Section~\ref{sec:discussion} regarding risks of using \GPLLM in regards to best practices of scientific investigations, present arguments for nonetheless pursuing evolving code with LLM-support, and suggest standards and open questions.
Finally, in Section~\ref{sec:conclusions--future} we conclude with a summary of the paper's contributions.

%% file: background.tex
\section{Background: Large Language Models}
\label{sec:background}

This section provides background on Large Language Models and code models. 
\subsection*{Large Language Models}\label{sec:LLMs}

Language models~(LM) generatively model the statistical likelihood of a corpus of text~\cite{liu2023pre}, implying they can generate text completions using pattern matching between prompt text and text seen during training.  This capability makes them extremely useful for natural language  tasks such as translation, summarization, or text classification. 
The capabilities of language models abruptly accelerated
with the adoption of transformer architectures~\cite{vaswani2017}.
Transformer architectures avoid the constraints of prior models implying that they had to be trained serially and now allow vast amounts of computation to be marshaled for training with parallelization. 
They also leverage training on the task of token sequence unmasking or completion which allows training on massive, public, (and free) unlabeled text corpuses such as the web pages of the Internet and digital libraries.
The``Large" in ``Large Language Models'' typically refers to language models with at least 10B parameters.  
LLMs catapulted to global attention and wide-spread adoption in 2022 with the introduction of OpenAI's ``GPT"  series (Generative Pre-trained Transformer,~\cite{radford2019language,
  brown2020language, openai2023gpt4}) which ranged from 120M to 175B parameters and used training sets that ranged in size from 1B~\cite{radford2019language} to 300B~\cite{brown2020language}. 
Chat-GPT made the biggest impact due it being released with a free and  easy-to-use question-answer interface. 
This interface (also provided programmatically via an API) accepts from some actor $A$,  a natural language text sequence, which functions to $A$ as a query, question or ``\textit{prompt}''.   
$A$ assigns knowledge, meaning, and intention to the text in the prompt.
They can provide problem context, problem solution examples, chain of thought reasoning, or other information, thought by them, to aid the LLM's generative pattern matching and completion. 
As output, the LLM provides a sequence of natural language  text, intended to be the ideal generative pattern-based completion to the prompt.
This input-output relation $f$ can be denoted simply as $f: \mathcal{T}^n \rightarrow \mathcal{T}^l, response = f(prompt | \Theta), n,l \in \mathbb{Z}^{+}$ where $\mathcal{T}^i$ denotes a sequence of text of length $i$ and $\Theta$ is the LLM parameters. 
It is very complicated to more specifically describe a transformer, let alone a GPT, e.g.~\cite{phuong2022formal}. Part~1 of one description, written in a 'pure mathematical' style runs 8 pages, see~\href{https://drive.google.com/file/d/1hqrHAhZAoDpFsnp1G0fQrQ1SxFtrOhEp/view}{this report}. Note that prompt text is tokenized prior to being input to the LLM so frequently LLM input is also described as tokens. 
All LLMs have a finite token capacity for a prompt-response pair. Often commercial LLMs charge by the token. 
An LLM remembers nothing between prompts. 
To link together an interdependent series of tasks, model responses, and even prior prompts, this information must be aggregated and included in subsequent prompts.


Despite many persuasive examples, LLMs exhibit a number of unresolved issues.
These include, for example,  the fact that LLMs offer \textit{no correctness guarantee}.   
In fact, because they are not retrieving information and instead are pattern-based, they have been documented to return confabulated references, facts and fallacious logic~\cite{ji2023hal}. This is often called ``hallucination''.  Approaches to address hallucination include human verification, automated verification, requesting the model to explain itself, and modifications to the RLHF layer. 
While a complete and robust successful approach is not yet available,  useful and timely progress on this challenge is occurring.

Second training and using an LLM is \textit{costly}~\cite{brown2020language}.  This cost scales with the number of model parameters and the size of the training data set.  
When deployed, due to how many parameters a model has, it is also costly to use it.  
This  expensive  footprint impedes  fair accessibility~\cite{strubell2020energy,patterson2021carbon,wu2022sustainable,kaack2022aligning}. 
As in the case of addressing hallucination, practical approaches are already in progress.

Third, while their human users, i.e. the actors $A$, no longer  have to resolve the highly technical or operational choices required to design and train deep neural networks,  they must alternatively redirect their focus on setting up the prompt(s) for the task they want the LLM to assist with or handle. In other words, they must focus on \textit{prompt engineering}.  
Prompt engineering must consider how to efficiently use the constrained token capacity of the prompt window. 
It must pack the problem context, historical information assisting with problem solving continuity, specifications of response format, and task  structure within the prompt's size limit.
This is more challenging to efficiently accomplish  if a series of prompts is needed while the LLM remembers nothing between prompts.
When prompt engineering must be programmed into a system, even more complicated design decisions have to be made. 
To date, mainly for human-model interactions, effective prompt engineering techniques have been developed.
One basic strategy of many of these approaches is to provide the LLM with
\textit{contextual information} that guides it to attend to a solution-appropriate pattern. 
Adding facts as context to the
 prompt is also effective to some extent~\cite{zhou2022teaching}.  
Other techniques include 
few-shot prompting~\cite{izacard2022few}, 
chain-of-thought~\cite{wei2022chain}, 
self-consistency~\cite{wang2023selfconsistency,shao2023synthetic}, "Tree-of-thoughts"~\cite{yao2023tree} and
 ``Ask-Correct''\cite{liventsev2023fully}. 

Fourth, LLMs may be negatively \textit{biased}. 
A seminal paper, on deep networks in general, explained how negative racial, gender, and other biases within a data set could end up being entwined with the functional capabilities of an LLM\cite{raji2020saving}.  
Negative biases and the general unpredictable nature of LLM behavior pose a safety issue.
With off-the-shelf LLMs, the training set is rarely shared and there are examples where training data has been used without permission\cite{appel2023generative} or clearly not reviewed. 
 Stricter filtering of training content and better transparency into a training sets data will provide better safety but  
bias-control is likely to be a longstanding open problem of LLMs.

Fifth,  prompts for one problem for one LLM should not be expected to transfer to another LLM.
Across LLM, their behavior is inconsistent. 
Additionally, LLMs that are publicly released with no cost should be treated much like Google's Search Engine with respect to the model makers not releasing any information about their version or updates. See, e.g.~\cite{chen2023chatgpt,du2023improving}.

Finally, and arguably most importantly, despite their ease of accepting and responding in natural language, LLMs are \textit{not capable of general human intelligence}. They are simply token-driven, token-outputting, generative pre-trained transformer models!  To emphasize this point, one can find a highly respected community of AI  researchers who study deep facets of general intelligence (artificial and human) and who work to correct inaccurate, overblown public perception that LLM's are capable of artificial general intelligence.  They show for example, that LLMs trained on “A is B” fail to learn “B is A” \cite{berglund2023reversal} and that LLMs cannot perform simple analogical reasoning~\cite{moskvichev2023conceptarc}
or mathematics~\cite{ding2023fluid}.  
Tom Dietterich clarifies that LLMs have "pointwise" understanding that allows them to provide appropriate responses to individual queries, while they lack “systematic” understanding— the ability to provide appropriate responses across an entire range of queries or situations.  He states ``When people complain that an AI system doesn’t ‘truly’ understand, I think they are often saying that while the system can correctly handle many questions/contexts, it fails on very similar questions/contexts. Such a system cannot be trusted to produce the right behavior, \textit{in general}.” (our italics)~\cite{mitchell2023,dietterich2023}.

\subsubsection*{Code models}\label{sec:code-models}
Of interest and relevance to genetic programming and other closely related communities such as search-based software engineering, LLMs can perform a number of types of coding tasks. 
Specific models for coding have been trained. 
For example, Meta released the Code Llama family of models, with LLMs specifically pre-trained and tuned for coding in August  2023 by Meta~\cite{roziere2023code}. 
They are free for research and commercial use and there are three models: 
Meta calls \textit{Code Llama} a foundational code model, 
\textit{Code Llama - Python} is specialized for Python;
and \textit{Code Llama - Instruct}, is fine-tuned for understanding natural language instructions about coding. 
These models come in different ``sizes'', i.e. with different number of parameters, with implications on how well they will work (the more parameters the better) and how much compute they use (fewer parameters use less compute).     
Meta claims that in its own benchmark testing, Code Llama outperformed state-of-the-art publicly available LLMs on code tasks~\cite{roziere2023code}.
One particular intersection between of the code-model community and Genetic Programming is benchmarks. 
Code Llama is evaluated with HumanEval, MBPP and APPS, as well as MultiPL-E and GSM8K~\cite{roziere2023code}.
Differentiating between research questions to pursued by the LLMs for coding community and research questions to be pursued by the GP community is an open challenge~\cite{o2020automatic}.



%% file: methods.tex
\section{A General \GPLLM Algorithm} \label{sec:spectrum}

In Section~\ref{sec:gpllm-algor-overv} we present a \GPLLM algorithm.  In Section~\ref{sec:gpllm-operators} we describe its
LLM based operators. Finally, in Section~\ref{sec:gpllm-prompts} we describe
prompts and prompt functions.

\subsection{\GPLLM Algorithm Overview}
\label{sec:gpllm-algor-overv}

\begin{algorithm}[htb]
  \footnotesize           
  \SetKwComment{Comment}{//}{}
  \SetKwInOut{Input}{Input}  \SetKwInOut{Output}{Return}
  \caption{\GPLLM \\Operators $i_{LLM}$, $e_{LLM}$, $\phi_{LLM}$, $s_{LLM}$, $v_{xoLLM}$, $v_{muLLM}$, $r_{LLM}$, $b_{LLM}$}
  \label{alg:gp_llm}
  \Input{\textbf{RUN HYPER-PARAMETERS}\\
$g$~: Generations, $n$~: Population size, \\
 $\LLM$~: LLM,\\
\textbf{PROMPT-FUNCTIONS (HYPER-PARAMETERS)}\\
    $\rho_i(\prims)$~: Prompt for initialization,\\
    $\rho_e(\mathbf{a})$~: Prompt for evaluation,\\
    $\rho_{\phi}(\mathbf{a})$~: Prompt for fitness measure,\\
    $\rho_s(\mathbf{a},)$~: Prompt for selection,\\
    $\rho_{v_{xo}}(\mathbf{a})$~: Prompt for crossover,\\
    $\rho_{v_{mu}}(\mathbf{a})$~: Prompt for mutation,\\
    $\rho_r(\mathbf{a})$~: Prompt for replacement,\\
    $\rho_b(\mathbf{a})$~: Prompt for picking best solution,
    }
  \Output{
    $P^*$~: best solution 
  }
  $P \gets i_{\LLM}(n, \rho_i \gets \rho_i(\prims))$ \Comment{Random initialization of population}\label{alg:gp_llm_init}
  \Comment{Iterate over generations}
  \For{$i \in [1, \dots, g]$}{ 
    $\mathbf{y} \gets e_{\LLM}(\rho_e \gets \rho_e(P, D, E))$ \Comment{Execute solution}\label{alg:gp_llm_execution}
    $\mathbf{f} \gets \phi_{\LLM}( \rho_{\phi} \gets \rho_{\phi}(\mathbf{y}, D))$ \Comment{Measure the fitness of the solution}\label{alg:gp_llm_fitness_measure}
    $P' \gets \emptyset$ \Comment{New population}
    \Comment{Iterate over population}
    \While{$|P'| \le n$}{ 
      $p_k, p_l \gets s_{\LLM}(P, \rho_s \gets \rho_s(P)))$ \Comment{Select two parents}\label{alg:gp_llm_selection}
      $p'_k, p'_l \gets v_{xo\LLM}( \rho_{v_{xo}} \gets \rho_{v_{xo}}(p_k, p_l)))$ \Comment{Variation with crossover} \label{alg:gp_llm_xo}
      \Comment{Iterate over children}
      \For{$j \in [1, \dots, n]$}{ 
        $p' \gets v_{mu\LLM}(\rho_{v_{mu}} \gets \rho_{v_{mu}}(\prims,p')))$ \Comment{Variation with mutation} \label{alg:gp_llm_mutation}
      }
      $P' \gets P' \cup \{p'_k, p'_l\}$ \Comment{Add to new population}
    }
    $P \gets r_{\LLM}( \rho_r \gets \rho_r(P')))$ \Comment{Replace the population} \label{alg:gp_llm_replacement}
  }
  \Return $P^* \gets \textnormal{b}_{\LLM}(P, \rho_b())$ \Comment{Best solution } \label{alg:gp_llm_rank_best}
\end{algorithm}

\GPLLM is described in Algorithm~\ref{alg:gp_llm}. 
 It is intentionally general -- each of its operators uses an LLM; while it is likely that many variants will use fewer. 
Algorithm~\ref{alg:gp_llm} exhibits a number of features that make it
quite dissimilar to GP.  
The first difference is that the unit of
evolution, code, is represented as a variable-length sequence of text in code syntax.
This sequence is
\textit{executable} but \textit{it is not a parse-tree}. 

All evolutionary operators in Algorithm~\ref{alg:gp_llm} (named in its caption)  are intended to use an LLM. They initialize a population of candidate genotypes (code, line~\ref{alg:gp_llm_init}), execute and evaluate them (lines~\ref{alg:gp_llm_execution} and \ref{alg:gp_llm_fitness_measure}), select parents (line~\ref{alg:gp_llm_selection}), create children with variation (lines~\ref{alg:gp_llm_xo} and \ref{alg:gp_llm_mutation}), and replace the old population with the old one (line~\ref{alg:gp_llm_replacement}).  Finally, Algorithm~\ref{alg:gp_llm} includes an LLM-based operator for designating the run's solution (line~\ref{alg:gp_llm_rank_best}). 

\subsection{LLM-based Operators,}\label{sec:gpllm-operators}
LLM-based operators are very different from well established evolutionary operators that are used in GP.  A \GPLLM operator takes three steps:
\vspace{12pt}

\begin{leftbubbles}
\begin{enumerate}
\item \texttt{Formulate} $\rho \gets \rho(\cdot)$ : Compose the prompt via calling the operator's prompt-function. 
\item \texttt{Interface} $r = f(\rho | \theta)$ : Send the prompt to the LLM and collect the LLM's response. $r \in \mathcal{T}^{m}$
\item \texttt{Check}  $r' = c(r, \cdot)$. Ensure $r$ is well formed.
\end{enumerate}
\end{leftbubbles}

%

\vspace{12pt}
Descriptions of LLM-based operators in the package \ALFAECLLM can be found in Appendix~\ref{sec:appdx-operators}.
A description of a LLM-based mutation operator from the package \ALFAECLLM follows:  
\vspace{12pt}
\begin{leftbubbles}
$V_{muLLM}$, \textbf{Mutation Operator}
\begin{enumerate}
\item \textbf{Formulate} prompt with instructions to alter one parent solution $p$, using elements from primitives.
\item \textbf{Interface} to the LLM to execute the prompt and collect the response.
\item \textbf{Check} response by extracting child solution from response. If it is not properly formatted, return the parent solution $p$.
\end{enumerate}
\end{leftbubbles}
\vspace{12pt}
As we observe in Section~\ref{sec:related-work}, not
every \GPLLM variant uses an LLM for every operator and
none use an LLM for code evaluation. LLM EA variants
which evolve types of solutions from a broader set than code,
e.g. evolve text, more frequently use an LLM base evaluation~(and
execution operator).

\subsection{Prompt-functions and Prompts}\label{sec:gpllm-prompts}

A new and significant parameter of each operator is its prompt,
$\rho$. \noindent A prompt \textit{variable or object} is a sequence
of text decomposed into elements:
\begin{leftbubbles}
\begin{quotation}
\noindent \texttt{<$\rho$> ::=  <EXAMPLES><QUERY><PRIMITIVES><RESPONSE\_FORMAT>}\\
$\rho$ is a sequence of text  $\rho \in \mathcal{T}^m$.
\end{quotation}
\end{leftbubbles}
\vspace{12pt}

Some prompts contain problem-dependent information.  
For example the prompts for initialization and mutation include the problem primitives. 
Some elements in a prompt need to be added at run-time. 
For example, solutions undergoing crossover or a population pool for selection.  
Run-time information is denoted in the template using \texttt{\{\}}s.
For example, a mutation prompt \texttt{<$\rho_{mu}$>} template for a symbolic regression problem is shown in
Figure~\ref{fig:mutation_prompt_ex}.

\begin{figure}[h]
\begin{leftbubbles}
{\scriptsize
<EXAMPLES> ::= \{n\_samples\} examples of mathematical expressions are: \{samples\}\\
<QUERY> ::= Rephrase the mathematical expression \{expression\} into a new mathematical expression.\\
<PRIMITIVES> ::= Use the listed symbols \{primitives\}.\\
<RESPONSE\_FORMAT> ::= Provide no additional text in response. Format output in JSON as
\{\{"new\_expression": "<expression>"\}\}
}
\end{leftbubbles}
\caption{Symbolic Regression mutation prompt template $\rho$}
\label{fig:mutation_prompt_ex}
\end{figure}

Because prompts may contain problem-dependent and run-time information, they must be created at run-time. This means that the \GPLLM must be provided with prompt-function  hyper-parameters (overloading the prompt variable's name $\rho$) that return these prompts. We generalize prompt-functions to use a list of arguments $\mathbf{a}$. 
The general signature of a prompt-function is $\rho \gets \rho(\mathbf{a})$.
The \ALFAECLLM  package  provides examples of specific signatures for a symbolic regression problem. One function for the mutation operator is shown in Figure~\ref{fig:ex_mutation_impl}. 
\begin{figure}[h]
\begin{lstlisting}
def form_prompt_rephrase_mutation(self, expression: str, samples: Optional[List[Any]]=None) -> str:
    if samples is not None:
        n_samples = min(len(samples), self.n_shots)
        // Randomly sample examples to provide context for the LLM
        sample_input = random.sample(list(samples.keys()), n_samples)
    else:
        sample_input = ""
        n_samples = 0

    prompt = self.REPHRASE_MUTATION_PROMPT_FEW_SHOT.format(
        expression=expression,
        constraints=self.constraints,
        samples=sample_input,
        n_samples=n_samples,
    )
    return prompt
\end{lstlisting}
\caption{\texttt{self.REPHRASE\_MUTATION\_PROMPT\_FEW\_SHOT} uses the prompt template in Figure~\ref{fig:mutation_prompt_ex}. It is a Python implementation of a function formulating a mutation prompt~($\rho_{mu}$). The arguments $\mathbf{a}$ for this function are \texttt{expression} and \texttt{samples}. }
\label{fig:ex_mutation_impl}
\end{figure}

\noindent This results in the prompt and response of Figure~\ref{fig:ex_mut_prompt}.

\begin{figure}[h]
\begin{rightbubbles}
{\scriptsize
2 examples of mathematical expressions are:
['((x0 + x1) * (x0 - x1) + 1)', 'x0 + x1 * (1 - 0)']

Rephrase the mathematical expression (x0 * x1) + (1 - 0) into a new mathematical expression.
Use the listed symbols ['*', '+', '-', 'x0', 'x1', '0', '1'].

Provide no additional text in response.
Format output in JSON as \{"new\_expression": "<new expression>"\}
}
\end{rightbubbles}
\begin{leftbubbles}
{\scriptsize
  \{"new\_expression": "(x0 * x1) + 1"\}
  }
\end{leftbubbles}

\caption{Example prompt and response for mutation.}
\label{fig:ex_mut_prompt}
\end{figure}

The response is then formatted, see Figure~\ref{fig:ex_mutation_resp}. This results in the response format as shown in Figure~\ref{fig:ex_mutation_resp}.
\begin{figure}[h]
\begin{lstlisting}
def format_response_rephrase_mutation(self, response: str, expression: str) -> str:
    try:
        phenotype = json.loads(response)["new_expression"]
    except (json.decoder.JSONDecodeError, KeyError, TypeError) as e:
        phenotype = expression
        logging.error(f"{e} when formatting response for rephrase mutation for {response}")

    return phenotype
\end{lstlisting}
\begin{leftbubbles}
  {\scriptsize
RESPONSE:
\{"new\_expression": "(x0 * x1) + 1"\}

INDIVIDUAL (PHENOTYPE):
(x0 * x1) + 1
  }
  \end{leftbubbles}
\caption{Example LLM mutation prompt response formatting python code implementation and output.}
\label{fig:ex_mutation_resp}
\end{figure}

%

Koza, in the online post~\cite{koza2023}
lays out 5 major preparatory steps to be followed ahead of a GP run. 
Steps~1~and~2 involve specifying the problem-dependent primitives, Step~3 involves specifying a problem-dependent fitness measure,  Step~4 involves specifying run parameters, and Step~5 involves designating the solution of the run\footnote{Koza left out specifying the (problem-dependent) input-output examples needed for candidate execution and fitness evaluation.}. The nature of the preparatory efforts of Steps~1,2, and 4 change radically with LLMs.  They culminate in a new set of at least $8$ hyper-parameters which are the prompt-functions, one per LLM operator, that each return a bespoke prompt, and which each use some set of prompt engineering techniques to express the purpose of the operator they serve.  Note the human expertise and effort required to prepare the prompts.
The three major preparatory steps ahead of a \GPLLM run, required of a person,  are:\\
\begin{leftbubbles}
SPECIFY:
\begin{enumerate}
\item the programming language that will express the candidate solutions plus problem-dependent hand-written primitives and any primitives built-in to the programming language to be used.
\item the prompt-functions of all operators implemented using an LLM. 
\item the hyper-parameters for controlling the run, including the termination criterion, i.e. Run Hyper-Parameters
\end{enumerate}
\end{leftbubbles}
\vspace{12pt}

\paragraph{Other Considerations}  Because Algorithm~\ref{alg:gp_llm} integrates an LLM,  the  run-time cost of interfacing with the LLM to execute a prompt and collect a response needs to be considered. It can be broken down into time and money.  An indirect cost of integrating the LLM is that of pre-training it. This is considerable and we defer discussion of how to consider the pre-training cost to~Section \ref{sec:discussion}. The change in the preparatory steps, particularly Step~2 where the prompt-functions must be designed, changes the nature of human effort toward preparing a GP run. 

Also  noteworthy is the difference in computational effort between GP and \GPLLM. We usually measure the computational effort of a GP run in terms of its cost-dominating operator - fitness evaluations. While fitness evaluations, likely external, i.e. not with LLM, are still integral to computational effort in \GPLLM, prompting and token counts also need to be counted.  When the token and prompt costs equal or exceed that of fitness evaluation, they should be incorporated into the algorithm's computational effort.

\input{related_work}

%% file: related_work.tex
\section{Related Work}
\label{sec:related-work}

In this section we present existing work at the intersection of EAs and LLMs, with a focus on  \GPLLM variants. 
This area is quite new so we include both peer-reviewed papers and non-peer reviewed papers found on \url{https://arxiv.org} at the time of writing.
We do not consider work which uses EAs to improve LLMs without using LLM-based operators, .e.g.~\cite{ling2023evolutionary} 

\subsection{By Problem Domain}\label{sec:rw-problemdomain}

We first group contributions by problem domain. 
We observe an array of problem domains that we group as 
\begin{inparaenum}[\itshape 1)]
\item  Code generation,
\item Neural Architecture Search, 
\item Game Design,
\item Prompt Generation.
\end{inparaenum}
Table~\ref{tab:rw_overview} identifies every contribution by its authors, citation in this article,
title, peer review status, and the problem domains it addresses.  We compare with each problem domain below.

\begin{table}[htb]
  \centering
  \footnotesize
  \caption{Overview of work at intersection of LLM and EA. PR indicates official peer-reviewed publication.}
  \label{tab:rw_overview}
  \begin{tabular}{l|p{5cm}|p{3cm}|l}
    \textbf{Author} & \textbf{Title} & \textbf{Problem} & \textbf{PR}\\
    \hline
    \multicolumn{4}{c}{\textbf{Code Evolution}} \\
      \hline
    \citeauthor{liventsev2023fully}~\cite{liventsev2023fully} & Fully autonomous programming with large language models.& Program synthesis & \checkmark\\
    \hline
    \citeauthor{zelikman2023selftaught}~\cite{zelikman2023selftaught} & Self-Taught Optimizer (STOP): Recursively Self-Improving Code Generation & Code for Optimization & \\
    \hline
    \citeauthor{lehman2022evolution}~\cite{lehman2022evolution} & Evolution through large models. &  Code for Agent controller & \\
    \hline
    \citeauthor{bradley2023}~\cite{bradley2023} & The openelm library: Leveraging progress in language models for novel evolutionary algorithms. &  Code for Agent controller,\newline Boolean Parity,\newline Program synthesis,\newline Text  &  \checkmark\\
    \hline
    \citeauthor{meyerson2023language}~\cite{meyerson2023language} & Language Model Crossover: Variation through Few-Shot Prompting & Code for Agent Controller,\newline Symbolic Regression,\newline Boolean Parity,\newline Text & \\    
    \hline
    \citeauthor{ma2023eureka}~\cite{ma2023eureka} & Eureka: Human-Level Reward Design via Coding Large Language Models & Code for reward function & \\
    \hline
    \citeauthor{chen2023evoprompting}~\cite{chen2023evoprompting} & Evoprompting: Language models for code-level neural architecture search.& Code for Neural Architecture Search & \\
    \hline
    \citeauthor{Nasir2023LLMaticNA}~\cite{Nasir2023LLMaticNA} & Neural architecture search via large language models and quality-diversity optimization. & Code for Neural Architecture Search & \\
    \hline
    \multicolumn{4}{c}{\textbf{Text Evolution}} \\
    \hline
    \citeauthor{guo2023connecting}~\cite{guo2023connecting} & Connecting Large Language Models with Evolutionary Algorithms Yields Powerful Prompt Optimizers & Prompt Search& \\
    \hline
    \citeauthor{fernando2023promptbreeder}~\cite{fernando2023promptbreeder} & Promptbreeder: Self-Referential Self-Improvement Via Prompt Evolution & Prompt Search& \\
    \hline
    \citeauthor{xu2023wizardlm}~\cite{xu2023wizardlm} & Wizardlm: Empowering large language models to follow complex instructions & Data for LLM tuning & \\
    \hline
    \citeauthor{lanzi2023chatgpt}~\cite{lanzi2023chatgpt} & Chatgpt and other large language models as evolutionary engines for online interactive collaborative game design. & Text for Game design & \checkmark\\
    \hline
    \citeauthor{sudhakaran2023mariogpt}~\cite{sudhakaran2023mariogpt} & MarioGPT: Open-Ended Text2Level Generation through Large Language Models & Text for Game design & \\
  \end{tabular}
\end{table}

\paragraph{1. Code Generation}
Within \textit{code  generation} are works 
evolving agent controllers or their reward functions~\cite{lehman2022evolution,ma2023eureka},
solving program synthesis, symbolic regression and Boolean parity problems~\cite{bradley2023,meyerson2023language,liventsev2023fully},
and generating meta-heuristics~\cite{zelikman2023selftaught}.

\citeauthor{lehman2022evolution}~\cite{lehman2022evolution} is the earliest paper with an evolutionary operator using an LLM.  Within a method called Evolution through Large Models~(ELM), it uses an LLM-based mutation operator and a Quality-Diversity technique called MAP-Elites to generate training examples that can be used to fine tune a LM for a particular context: game terrain.   
EUREKA~\cite{ma2023eureka} performs LLM-based variation for the evolution of a reward function for agent-based Reinforcement Learning. The approach uses an LLM create reward function variations based on reflective summarizations of the agent performance. It also allows human interaction and feedback on the reward function.

Attempts to solve the GP community's  Program Synthesis Benchmark
2~\cite{helmuth2022applying} with LLMs tend to generate
programs that semantically resemble the correct answer but that have subtle flaws. 
\citeauthor{liventsev2023fully}~\cite{liventsev2023fully} introduce a unique and startling effective evolutionary approach to program synthesis called SEIDR: Synthesize, Execute, Instruct, Debug and
Rank. 
First, program snippet solutions are LLM-generated~(Synthesize). Second, with extra program context, every snippet is externally executed and assigned a fitness~(Execution). Next, some solutions are selected based on a top-$k$ ranking~(Rank). A repair phase ~(Instruct) then tries to analyze these failed solutions by considering their performance~(Debug). 
To balance exploitation (repairing current solutions) and exploration (replacing the current solution), a beam-search algorithm is used. 
The SEIDIR framework outperforms Codex highlighting  the power of LLM operators and evolutionary approaches. On Python and C++ on
the PSB2 benchmark~\cite{helmuth2022applying} SEIDR outperforms the
PushGP baseline and achieves the state-of-the-art result with 19
solved problems out of 25 with under 1000 program executions.
While SEIDR is compared to GP in ~\cite{liventsev2023fully}, only a simple GP variant is compared and the pre-training effort underlying the LLM is not considered in the comparison.

A variety of typical simple GP problems~(e.g. Simple agent controllers, Symbolic Regression, OneMax) are investigated
by \citeauthor{bradley2023}~\cite{bradley2023}. 
They introduce OpenELM, an open-source Python library for designing evolutionary
algorithms that leverage LLMs to generate variation, as well as to assess fitness and measures of diversity. 
Many of the same problems are investigated by \citeauthor{meyerson2023language}~\cite{meyerson2023language}. 
They input text-based genotypes that are either code, plain-text sentences, or equations to the LLM. 
They then interpret and use the corresponding LLM responses as those genotypes’ offspring. 
Their experiments highlight the versatility of LLM-based crossover and the success of the LLM-based approach.

Finally, a meta-heuristic approach that is not explicitly described as an EA, but that is similar in spirit, and which uses a LLM is introduced by
\citeauthor{zelikman2023selftaught}~\cite{zelikman2023selftaught}. An
LLM is used to generate a scaffolding program that is iteratively improved. 

\paragraph{2. Neural architecture search~(NAS)} NAS using a Python code
representation is investigated by
\citeauthor{chen2023evoprompting}~\cite{chen2023evoprompting}. They
use LLM-based mutation and crossover operators.  
For prompt engineering they use a few shot learning technique and they
fine tune with  the sub-population that was not selected. 
In other NAS work
\citeauthor{Nasir2023LLMaticNA}~\cite{Nasir2023LLMaticNA} use
an LLM and a Quality-Diversity algorithm to obtain variations of Python code defining a neural architecture.
Their system LLMatic create diverse networks that are high-performing. 

\paragraph{3. Game Design}
Game design is another problem domain where EA and LLM operators have
been combined. Typically the genotype representation is text. 
\citeauthor{lanzi2023chatgpt}~\cite{lanzi2023chatgpt}
present a collaborative game design framework that combines
interactive evolution and LLMs to simulate the typical human design
process. They use interactive evolution for selection and LLMs
for a recombination and variation of ideas. Another
game design example is from
\citeauthor{sudhakaran2023mariogpt}~\cite{sudhakaran2023mariogpt}. They
use a fine-tuned GPT2 model to generate tile-based game
levels. It is used for LLM based initialization and variation of
prompt text that describes a game level.

\paragraph{4. Prompt Generation}
Hand-crafted prompts are often sub-optimal. 
Promptbreeder~\cite{fernando2023promptbreeder} mutates a
population of text-based ask-prompts, evaluates them for fitness on a training
set, and repeats this process over multiple generations to evolve
task-prompts. In a self-adaptive way, Promptbreeder uses its mutation-prompts to improve the task-prompts. 
Another work on discrete prompt optimization uses LLM-based evolutionary operators
and Differential Evolution to improve a population based on
the development set~\cite{guo2023connecting}. Finally, \citeauthor{xu2023wizardlm}~\cite{xu2023wizardlm}
starts with an initial set of instructions that they evolve and
instruct the LLM to rewrite step by step into more complex instructions.
The frequency of papers evolving
prompts is likely due to  the non-trivial nature of efficient
prompt construction and the easy means with which an LLM handles text.

\subsection{By Operators, Genotype and LLM designs}\label{sec:rw-rest}
Next, across the works, we consider which LLM-based evolutionary operators are used, what genotype representations are evolved, and LLM-specific design decisions.   LLM-specific design decisions fall into two classes: guiding the behavior of the LLM, and Prompt Engineering Techniques.

We observe LLM behavior being guided in two ways: 
\begin{asparadesc}
\item [Raising Model Temperature] LLMs have a temperature parameter. 
A low  temperature makes prompt completion more deterministic. 
Raising the temperature leads to more variability in prompt completion. 
Some approaches raise the LLM temperature to obtain more diverse solutions.
\item [Model Fine tuning] Update the LLM parameters based on some data and training procedure,
  $\Theta' = g(X, Y, \Theta)$. Note that when training data is generated by the model and the updated and fine-tuned model is used in an evolutionary run, the algorithm is self-adapting. 
\end{asparadesc}

We observe selections from the following set of Prompt Engineering Techniques, (see also Section~\ref{sec:background}). 
\begin{asparadesc}
\item [Zero-Shot] A simple predefined prompt $\rho = T, T \in
  \mathcal{T}$
\item [Template] The prompt is a template that is expanded at runtime
  using run-time information, $\rho = \rho(\mathbf{x})$
\item [Few-shot] An extension of Template providing examples of correct responses, $\rho =
  \rho(\mathbf{x}, X, Y), |X| = |Y|, X, Y \in \mathcal{T}$. Note, this
  is some times called in-context learning.
\item [Chaining] 
 A sequence of LLM calls, $y_i =
  f(\rho_{i-1}(y_{i-1})), y_0 = f(\rho_0()), i \in \mathbb{Z}^+$. For example,
  chain-of-thought is a type of chaining that does not presume any
  external environment changes~\cite{yao2023react}.
\item [Summarization] A combination of Template and Chaining~($i=2$)
  where an individual is first summarized and then provided as input
  to a template.
\item [Human Interaction] A human interacts with the LLM to  manipulate the prompts and responses, $\rho', y' = H(\rho,
  f(\rho|\Theta))$
\item [Optimization] A prompt's content is (externally)
  optimized according to some function $u$, $\rho' =
  g(\rho, u(f(\rho|\Theta))), u: \mathcal{T} \rightarrow \mathbb{R}, v
  = u(\rho)$
\end{asparadesc}

All papers use the LLM generative capabilities for variation, mutation and crossover.
A majority also use it for solution initialization, and a minority use it to measure fitness. 
This includes a case where the diversity of solutions is measured through string embedding
similarity.  
We observe that a wide variety of prompt
engineering techniques are used. The most popular technique is Template use.
This seems like an obvious approach to integrating run-time information such as genotypes and fitness scores.
Second is raising the temperature which increases solution diversity. In order of popularity, are:
\begin{inparaenum}[\itshape 1)]
\item Template~(8)
\item Changing Temperature~(3)
\item Chaining~(2)
\item Human Interaction~(2)
\item Few-shot~(2)
\item Zero-shot~(2)
\item Summarization~(2)
\item Optimization~(1).
\end{inparaenum}
Table~\ref{tab:rw} summarizes. 

\begin{table}[h]
  \centering
  \footnotesize
  \caption{Details of LLM operators, genotype representation and
    prompt engineering for related work on LLMs and EA. Use and prompting refers to LLM model manipulation and prompt engineering techniques.}
  \label{tab:rw}
  \begin{tabular}{l|p{3cm}|lll|p{3cm}}    
     &  & \multicolumn{3}{c}{\textbf{Genotype}} &  \\
    \textbf{Cite Nr} & \textbf{LLM Operators} & \textbf{Code} & \textbf{Text} & \textbf{Bits} & \textbf{Use \& Prompting} \\
    \hline
    \multicolumn{6}{c}{\textbf{Code Evolution}} \\
    \hline
    \cite{liventsev2023fully} & Mutation,\newline Initialization & \checkmark & &  & Template,\newline Changing temperature,\newline Chaining,\newline Summarization\\
    \hline
    \cite{zelikman2023selftaught} & Mutation,\newline Fitness Measure & \checkmark & & & Optimization \\ 
    \hline
    \cite{lehman2022evolution} & Mutation & \checkmark & & & Template,\newline Fine Tuning \\
    \hline
    \cite{bradley2023} & Mutation,\newline Initialization,\newline Crossover,\newline Fitness Measure~(QD) & \checkmark & \checkmark & \checkmark  & Template \\
    \hline
    \cite{meyerson2023language} & Mutation,\newline Initialization,\newline Crossover,\newline Fitness Measure~(QD) & \checkmark & \checkmark & \checkmark & Few-shot  \\    
    \hline
    \cite{ma2023eureka} & Mutation & \checkmark & & & Template,\newline Summarization,\newline Human interaction\\
    \hline
    \cite{chen2023evoprompting} & Mutation,\newline Initialization,\newline Crossover & \checkmark & \checkmark & & Fine tuning,\newline Few-shot,\newline Changing temperature \\
    \hline
    \cite{Nasir2023LLMaticNA} & Mutation,\newline Initialization & \checkmark & & & Zero Shot,\newline Changing Temperature  \\
    \hline
    \multicolumn{6}{c}{\textbf{Text Evolution}} \\    
    \hline
    \cite{guo2023connecting} & Initialization,\newline Mutation, \newline Crossover & & \checkmark & & Template\\
    \hline
    \cite{fernando2023promptbreeder} & Initialization,\newline Mutation & & \checkmark & & Template,\newline Chaining \\
    \hline
    \cite{xu2023wizardlm} & Mutation & & \checkmark & & Template\\    
    \hline
    \cite{lanzi2023chatgpt} & Mutation,\newline Initialization,\newline Crossover & & \checkmark &  & Template,\newline Human interaction\\
    \hline
    \cite{sudhakaran2023mariogpt} & Mutation,\newline Initialization,\newline Crossover & & \checkmark & & Zero Shot \\
  \end{tabular}
\end{table}

%% file: experiments.tex
\section{Demonstration of a simple \GPLLM variant}
\label{sec:experiments}


In Section~\ref{sec:an-gpllm-impl} we describe an implementation of
\GPLLM within a package and code.  In
Section~\ref{sec:experimental_setup} we describe an experimental
setup. In Section~\ref{sec:results} we describe, analyze and discuss
experimental results.

\subsection{\GPLLM implementation}
\label{sec:an-gpllm-impl}

Section~\ref{sec:alfaecllm-design} describes key\GPLLM elements within the package. 
Section~\ref{sec:alfaecllm-components} describes the package's key\GPLLM components.

\subsubsection{The \ALFAECLLM Package}
\label{sec:alfaecllm-design}

\ALFAECLLM design priority is simplicity and readability in order to
help a broad range of learners.  It is implemented as a
\texttt{Python} package of three modules: \textbf{Algorithms}, \textbf{Problem
Environments} and \textbf{Utilities}.
\begin{asparadesc}
\item [Algorithms] includes three algorithms, each in a separate file,
  \texttt{Evolutionary\_Algorithm}, \texttt{Tutorial\_GP} and
  \texttt{Tutorial\_LLM\_GP}. Each algorithm is in a file that imports
  common data and EA operators.
\item [Problem Environments] includes prompt-functions and strings for the prompt
  template of each LLM-based operator for an example use case -
  Simplified Symbolic Regression.
\item [Utilities] includes utilities for interfacing with LLMs and running
  and analyzing experiments.
\end{asparadesc}

\subsubsection{\ALFAECLLM Design}
\label{sec:alfaecllm-components}

Key design features of \GPLLM are:
\begin{asparadesc}
\item Its \textbf{evolutionary unit }(field \texttt{genotype} of structure \texttt{Individual}) is a symbolic expression for Symbolic Regression
\item Its \textbf{Prompt Engineering} uses templates and few-shot learning. 
\item \textbf{LLM API}, the \texttt{OpenAIInterface}, is a class that interfaces
  with OpenAI's GPT-3.5-turbo model. It relies on code provided in
  OpenAI cookbooks for interacting efficiently with the web API, see
  Figure~\ref{fig:ex_llm_api}. It also records the input and output to
  the API. In addition, it tries to reconnect with exponential back-off
  when exceptions are thrown from the API.

\item  \textbf{Extra Error handling} for LLM timeouts or incorrectly formatted
  responses. Exceptions from LLM operations are stored and as a
  fall-back the default LLM operator behavior is
  executed. E.g. default phenotype, fitness, random selection and no
  variation.
\item \textbf{Extra Logging} \texttt{generation\_history} stores each LLM API call
  and response, as well as statistics regarding number of tokens and
  response time. These are essential for debugging.
\end{asparadesc}

\begin{figure}[h]
\begin{lstlisting}
// Wrap call in a function that retries with exponential back-off    
@retry_with_exponential_backoff
def predict_text_logged(self, prompt: str, temp: float=0.8) -> Dict[str, Any]:
    n_prompt_tokens = 0
    n_completion_tokens = 0
    start_query = time.perf_counter()
    content = "-1"

    message = [{"role": "user", "content": prompt}]
    // Get response from gpt-3.5-turbo
    response = openai.ChatCompletion.create(
        model="gpt-3.5-turbo", messages=message, temperature=temp
    )
    // Logging information 
    n_prompt_tokens = response["usage"]["prompt_tokens"]
    n_completion_tokens = response["usage"]["completion_tokens"]
    content = response["choices"][0]["message"]["content"]
    end_query = time.perf_counter()
    response_time = end_query - start_query
    return {
        "prompt": prompt,
        "content": content,
        "n_prompt_tokens": n_prompt_tokens,
        "n_completion_tokens": n_completion_tokens,
        "response_time": response_time,
    }
\end{lstlisting}
\caption{Example LLM API implementation in Python.}
\label{fig:ex_llm_api}
\end{figure}

\subsection{Setup}
\label{sec:experimental_setup}

Experimental resources are listed in Table~\ref{tab:experiment_resources}.
\begin{table}[h]
  \centering
  \caption{Experiment resource descriptions.}
  \label{tab:experiment_resources}
  \begin{tabular}{ll}
    \textbf{Resource} & \textbf{Description} \\
    \hline
    Operating system & Ubuntu 22.04 LTS \\
    RAM & 64GB \\
    CPU & Intel i7-8700K 3.70GHz \\
    \hline
    Budget & 50 USD \\
    \hline
    Max runtime & $60000$ seconds \\
    \hline
    Fitness Evaluations~(FE) & $300$ \\
    \hline
    LLM version & \texttt{gpt-3.5-turbo-0613} \\
    Token max size~($\mathcal{T}^n$) & 4,096 \\
  \end{tabular}
\end{table}

For baseline algorithms, we include random GP-like explicit generation of solutions, Random, and Tutorial GP from the package.
We also directly prompt the LLM to generate random solutions, a method we call LLM.
We explore two \GPLLM variants.
One, \GPLLM, uses a LLM in all its evolutionary operators except its fitness measure. 
The second, \GPSomeLLM, only uses a LLM in its initialization, crossover, and mutation operators.   

%
%
%
%

We use the experimental parameters listed in Table~\ref{tab:experiment_settings}. 
\begin{table}[h]
  \centering
  \caption{Experiment settings. Note the limits
    to population size is due to the LLM input and output buffer size,
    $\mathcal{T}^{n,m}$. Limit to generations is due to LLM query time
    and budget. LLM operators use Few Shot examples for prompt engineering.}
  \label{tab:experiment_settings}
  \begin{tabular}{l|l|l|l|}
    \textbf{Parameter} & \textbf{\GP} & \textbf{\GPSomeLLM} & \textbf{\GPLLM} \\
    \hline
    Runs & \multicolumn{3}{|c|}{30}\\
    Crossover probability & \multicolumn{3}{|c|}{$0.8$}\\
    Mutation probability & \multicolumn{3}{|c|}{$0.2$}\\
    Population size &  \multicolumn{3}{|c|}{10}\\
    Generations &  \multicolumn{3}{|c|}{30}\\
    \hline
    Primitives & \multicolumn{3}{|c|}{+,-,*,\, $x_0, x_1, 1, 0$}\\
    Solution & \multicolumn{3}{|c|}{$x_0^2 + x_1^2$}\\
    Exemplar splits & \multicolumn{3}{|c|}{0.2 Hold-out, (0.7 Training, 0.3 Testing)}\\
    \hline
    Exemplars & 121 & \multicolumn{2}{|c|}{10}\\    
    \hline
    Few shot exemplars & NA & \multicolumn{2}{|c|}{2} \\
    Mutation & Subtree & \multicolumn{2}{|c|}{See Appendix~\ref{sec:prompts}} \\
    Crossover & Subtree & \multicolumn{2}{|c|}{See Appendix~\ref{sec:prompts}} \\
    Initialization & Ramped-Half-Half & \multicolumn{2}{|c|}{See Appendix~\ref{sec:prompts}} \\
    Max Depth & 5 & \multicolumn{2}{|c|}{NA} \\
    \hline
    Selection & \multicolumn{2}{|c|}{Tournament} & See Appendix~\ref{sec:prompts} \\
    Tournament size & \multicolumn{2}{|c|}{2} & NA\\
    Replacement & \multicolumn{2}{|c|}{Generational} & See Appendix~\ref{sec:prompts} \\
    Elite size & \multicolumn{2}{|c|}{1} & NA\\
    
  \end{tabular}
\end{table}

\subsection{Analysis}
\label{sec:results}

This section analyzes the demonstration. Section~\ref{sec:analysis}
analyzes run duration and cost. Section~\ref{sec:analys-over-gener}
analyzes solution size and runtime. Section~\ref{sec:llm-oper-analys}
analyzes LLM usage. Section~\ref{sec:llm-operation-errors} analyzes
LLM operation errors.

\subsubsection{Time \& Cost Analysis}
\label{sec:analysis}

Results for a run is in Table~\ref{tab:results_main}. Each run used
300 FEs. LLM operators are orders of magnitude slower and costlier
than \GP. \GPLLM takes the longest. \GPSomeLLM is slightly faster due
to fewer LLM calls. As expected \GPSomeLLM takes less time than
\GPLLM, due to fewer LLM calls by not using LLM for selection and
replacement. Note that \GP is faster than random due to the caching of
fitness evaluations.

{\scriptsize 
\begin{table}[h]
\caption{Cost and runtime~(seconds) results for compared methods all solving Simple
  Symbolic Regression. Average over 30 runs.}
\label{tab:results_main}
\begin{tabular}{lrrr}
\toprule
Name & Mean Duration (seconds) & STDEV  & Cost (USD)\\
\midrule
LLM & 837.16 & 416.12 & 2.63 \\
\GPLLM & 1664.30 & 1033.97 & 3.90 \\
\GPSomeLLM & 743.31 & 508.70 & 1.87 \\
\hline
\GP & 0.10 & 0.08 & 0.00 \\
\hline
Random & 0.18 & 0.01 & 0.00 \\
\bottomrule
\end{tabular}
\end{table}
}

\subsubsection{Size Analysis}
\label{sec:analys-over-gener}

\begin{figure}
  \centering
  \begin{subfigure}[b]{0.89\textwidth}
    \centering
    \includegraphics[width=\textwidth]{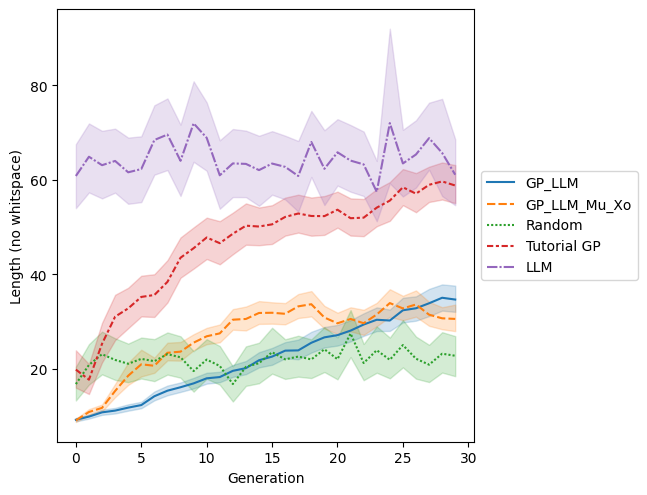}
    \caption{Simple Symbolic Regression solution sizes over generations.}
    \label{fig:size_generations}
  \end{subfigure}
  \begin{subfigure}[b]{0.89\textwidth}
    \centering
    \includegraphics[width=\textwidth]{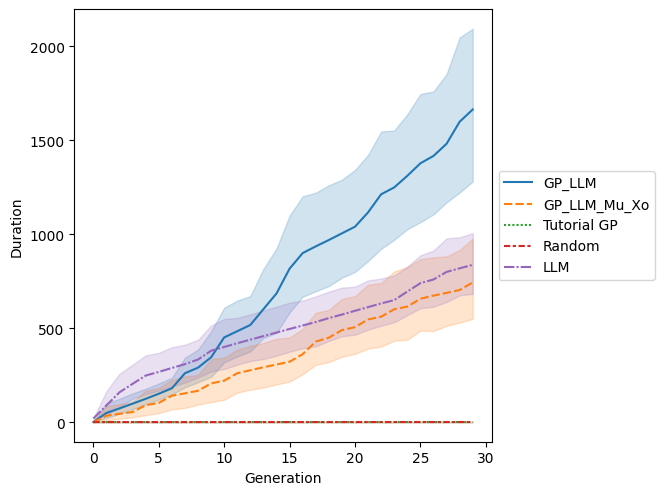}
    \caption{Duration~(s) of each generation.}
    \label{fig:duration_generations}
  \end{subfigure}
  \caption{Solution size and duration per generations}
  \label{fig:duration_and_suze}
\end{figure}

Figure~\ref{fig:size_generations} shows average solution size over
generations of a run. For \GPLLM mean solution size increases. With
\GPSomeLLM mean size increase up to generation 15 and then it
stabilizes. \GP fluctuates and the size is larger. Note, that there is
a solution simplification step for the representation for the LLM
based methods, but not for \GP.  \texttt{LLM only} has the longest solutions.

Figure~\ref{fig:duration_generations} shows duration of each
generation~(final generation is the total duration of the
experiment). \GPLLM takes the longest, as expected given that it has
the most calls to the LLM API. Note that the LLM execution time
includes API service restrictions and networking
limitations. \GPSomeLLM has fewer LLM calls, thus has shorter runtime
than \GPLLM. The GP runtime is several orders of magnitude lower.

\subsubsection{LLM Usage Analysis}
\label{sec:llm-oper-analys}

\begin{figure}
     \centering
     \begin{subfigure}[b]{0.79\textwidth}
         \centering
         \includegraphics[width=\textwidth]{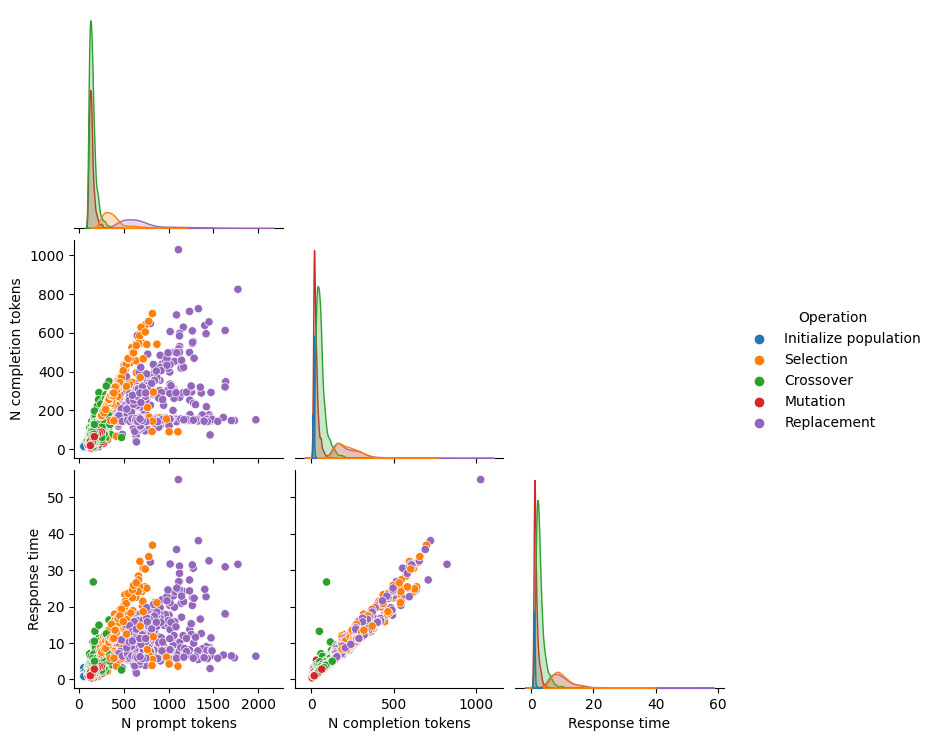}
         \caption{\GPLLM}
         \label{fig:llm_operations_gp_plus_llm}
     \end{subfigure}
     \begin{subfigure}[b]{0.79\textwidth}
         \centering
         \includegraphics[width=\textwidth]{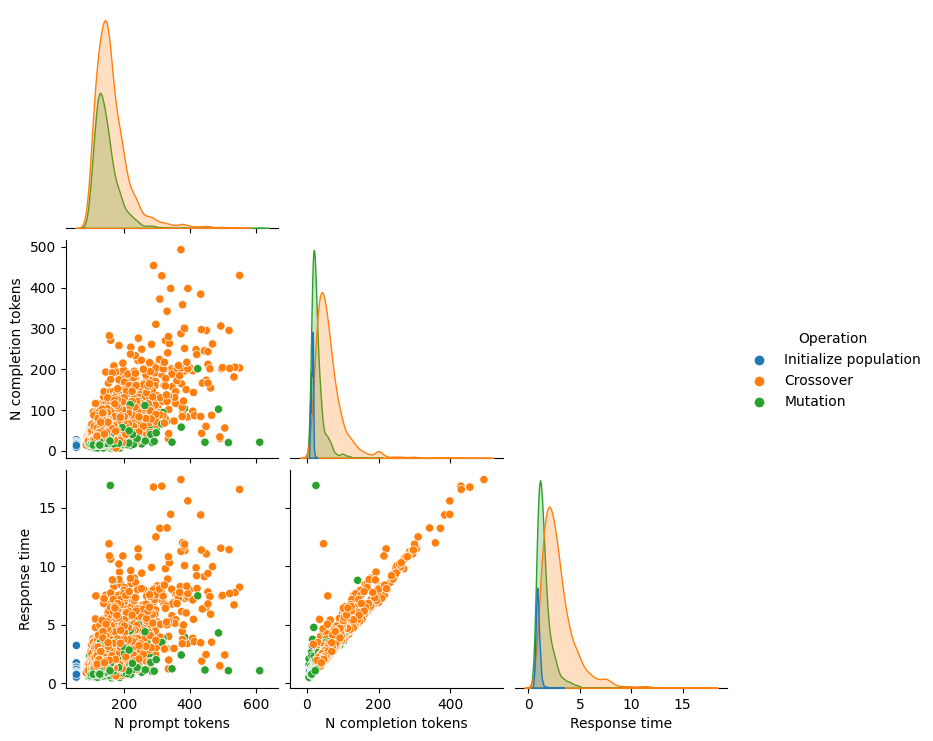}
         \caption{\GPSomeLLM}
         \label{fig:llm_operations_gp_plus_some_llm}
     \end{subfigure}
     \caption{LLM operation scatter plots and histograms of number of
       tokens in query, response~(Completion tokens) and response
       time. Color indicates the LLM operation}
     \label{fig:llm_operations}
\end{figure}

Figure~\ref{fig:llm_operations} shows statistics on LLM usage. We
observe that \GPLLM, see Figure~\ref{fig:llm_operations_gp_plus_llm},
has the largest number of prompt tokens, completion tokens and
response time. As expected, the LLM based operators for selection
and replacement use the most tokens and LLM time.

Initialization, mutation and crossover use fewer tokens and LLM
time. In Figure~\ref{fig:llm_operations_gp_plus_some_llm} we can more
clearly see that these LLM operators also behave as expected regarding
number of prompt tokens, number of completion tokens and response
time, i.e. initialization has shortest prompt~(it asks for an
individual) and crossover prompt contains examples and two parents~(it
asks for two children). Note, the observed linear relationship between
completion tokens and response time might be an artifact from the LLM
API service.

\subsubsection{LLM Operation Errors}
\label{sec:llm-operation-errors}

\begin{figure}
     \centering
     \includegraphics[width=\textwidth]{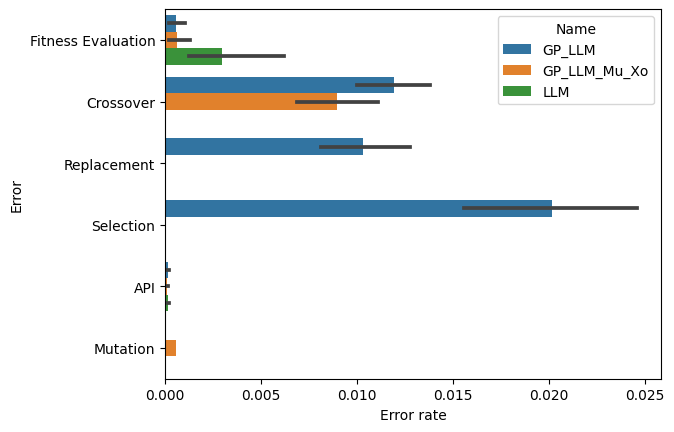}
     \caption{Error types for each method. X-axis error rate. Y-axis is the error type.}
     \label{fig:error_counts}
\end{figure}

Prompts are important, as well as the LLM that use the prompts for
inference. We observe that some LLM based operators generate fewer
errors than others. Figure~\ref{fig:error_counts} shows the error rate
for the different variants and operations. The highest error rate is
for selection. Replacement also throw errors. Both of these are most
likely due to the prompt size. The crossover also throws errors, this
can be from the requirement that the symbols of the parents should be
reduced. Note we would expect even more errors if we tried to enforce
the standard GP subtree crossover constraint of subtree swaps as well.

The fitness evaluation errors are due to malformed expressions from the LLM that the response formatting did not catch. 
Mutation results in very few errors. 
Initialization results in no errors, so this was a robust LLM operator. 
There are also errors from the LLM API service. 

%% file: discussion.tex
\section{Discussion}
\label{sec:discussion}

In Section~\ref{sec:discussion:comparison} we compare \GPLLM
and GP. In Section~\ref{sec:discussion:risks} we discuss risks and
other implications of using an LLM specifically to evolve code. In
Section~\ref{sec:discussion:why} we discuss why  \GPLLM
investigations should continue despite the risks. 
In Section~\ref{sec:discussion:conduct} we suggest
 guidance for conducting \GPLLM investigations. Finally, in
Section~\ref{sec:discussion:OpenRQs} we state some \GPLLM research directions.


\subsection{\GPLLM vs GP}\label{sec:discussion:comparison}

We summarize GP and \GPLLM differences in Table~\ref{tab:comparison_properties}.
Both GP and \GPLLM are EAs where the evolving unit is code. 
Both execute with the same procedural logic and operators of an EA.
Both represent code with  symbols naming problem primitives. 
But GP relies upon the code being represented by a structure which allows its execution and explicit,  manipulation through variation operators that enforce syntactic correctness. 
For example, Koza-style GP works with code in a parse tree structure, it executes the tree structure with an internal code interpreter, and it uses sub-tree exchange for crossover. 
In contrast \GPLLM works with a code snippet that is tokenized and acceptable for input to a LLM.  
It is able, if desired, to directly encode the snippet using a common programming language 
While it could task the LLM to execute the code snippet, it very reasonably generally uses an external interpreter.
Unlike GP, it forgoes explicit structural manipulation of the code during variation. 
Instead, it tasks the LLM to recombine, or mutate code by using a prompt, and the only  control it exercises over how the operation is accomplished is through the prompt formulation. 
The LLM -- a complex, pre-trained, pattern-based sequence completion system, that almost deceptively seems to understand language meaning while it truly does not, is effectively otherwise a black box~\cite{webson-pavlick-2022-prompt,lipkin2023evaluating}. 


\begin{table}[h]
  \footnotesize
  \centering
  \caption{Comparison of GP and \GPLLM.}
  \label{tab:comparison_properties}
  \begin{tabular}{p{3cm}|p{4.5cm}|p{4.5cm}}
    \textbf{Basis of Comparison} & \textbf{GP} & \textbf{\GPLLM} \\
    \hline
    Computational model/environment referenced by the code-evolving system  &
    Program execution model / environment &
    Program execution model/environment and LLM which is a generative pattern completion system using token/sequence-based pattern-matching with built-in patterns \\
    \hline
    Run of a code-evolving system &
    A GP run executes procedural software where the code is data, the operators work on code structure, and the code is bespoke evaluated and assigned numerical fitness. &
    A \GPLLM run executes procedural software that, among other things, composes text-based NL prompts, sends them as inputs to an LLM , and collects responses. (Prompts and responses at lowest level of description are sequences of tokens) \\
    \hline
    Code as desired solution (genotype-phenotype duality) &
    Genotype/phenotype is a data structure with structural properties, e.g. tree, list, stack, and executability &
    Code is token sequence with code-snippet meaning, it has no structural properties, and it has implicit pattern-related properties related to the patterns, patter—matching and bias within the LLM \\
    \hline
    Evolutionary Variation &
    Structural, blind to meaning &
    Not structural, blind to user beyond prompt content. Internal to LLM it is based on built-in patterns and is a black box. \\
    \hline
    Evolutionary Selection/Replacement &
    Comparative, based on numeric ranking and fitness represented as a number &
    Comparative, prompt could include fitness, could task LLM to rank, could include other bases of comparison. Blind to use beyond prompt content. Internal to LLM it is based on built-in patterns and is a black box. \\
    \hline
    Code evaluation &
    Uses bespoke execution environment (supporting run’s primitives) on top of a general-purpose program execution environment &
    Practical implementations will use a general-purpose program execution environment \\
    \hline
    Code Fitness &
    Numeric-based &
    Numeric or expressed with natural language \\
    
  \end{tabular}
\end{table}

Regardless of the LLM being a black box, when prompted with a variation operators' task, it can demonstrate startling capabilities that allow it to potentially respond, on average, with a code snippet that is a better adaptation than adaptations the equivalent GP operator generates.   
When these capabilities arise, a \GPLLM run is going to solve a problem with fewer fitness evaluations than GP.
The power of the LLM in this variation task capacity has been demonstrated, e.g. with a \GPLLM variant solving more of the GP community's program synthesis benchmarks~\cite{liventsev2023fully}.
The same holds for the \GPLLM's selection operator. 
In GP, selection depends on a numerical comparison of fitness. 
In \GPLLM, selection need not follow GP's approach. 
Selection depends on a prompt that contains the set of selection candidates, other information about them (including fitness). The prompt directs a selection of some subset, based on a criteria expressed in natural language. 
Then, again, the black box, i.e. the LLM,  is used to obtain a response and the \GPLLM algorithm logic has no further control.  
If and when the LLM, across many operator calls, selects more optimally than a GP selection operator, the \GPLLM run could converge to a solution with fewer fitness evaluations than that of a GP run.  


\subsection{Risks and other Implications of Using an LLM}\label{sec:discussion:risks}
We have just pointed out the obvious risk posed by \GPLLM's operators: the operator  (and human designer) surrenders its explicit control over the details of its behavior. 
\GPLLM also faces other risks and challenges(some generally raised in Section~\ref{sec:background}):



\begin{itemize} 
\item it seems that LLMs handle software engineering and coding better than facets of general intelligence, but the reason why is not known. This conundrum adds to the black box nature of the LLM.

\item 
an algorithm's success depends on prompt composition, while the sensitivity of an LLM to a prompt's composition is unreliable. 
Sensitivity would have to be probed and quantified, for each LLM independently.
Further, LLMs lack many facets of general intelligence, while to some degree, they deceptively appear to understand prompts. This leads to the risk of mistakenly assuming understanding when it does not exist.

\item 
all LLMs are fine-tuned to be aligned with goals. They display this bias. Neither evolution or coding are specific with any LLM's goals, but  biases relating to evolution and coding emerge from pre-training.  Some LLMs are specifically trained for coding: see LLama~\cite{roziere2023code} family, but may not be specifically trained on evolution. The biases of these models are poorly characterized or understood. 
\item 
success also depends on the design (and aforementioned bias) of the LLM. A LLM  is concisely described as a set of weights ($\theta_{LLM}$) that have been pre-trained under design decisions of neural network architectures, training algorithms, and data sets.  
Precisely predicting the combined impact of all of these design decisions is not practically feasible because an LLM is probabilistic and generative. 
A LLM can be tested and its test results released to describe its capabilities, but there is no description of its precise behavior with new prompts and tasks. 
\item 
LLM pre-training is often out of the hands of LLM users and data sets for training and testing are frequently not well documented or openly shared.
A researcher may not be able to ensure that the rote solution (and problem description) to the problem at hand is not within the training data upon which the model has pre-trained. 
\item 
when an LLM is used via a model-provider's API, experimental replicability is largely impossible.
\end{itemize}

Collectively these risks directly and significantly impact the quality of the science that can be conducted with a \GPLLM system. To cap this off, working with LLMs is non-trivial and resource intensive. This prompts a question $\dots$.

\subsection{Why conduct \GPLLM investigations?}\label{sec:discussion:why}
It is arguable that, despite a LLM-based EA system being less than ideal for scientific purposes,  reports of investigations should be welcome and accepted for publication.
A fundamental reason is that an LLM evolves code by drawing upon its pre-training on  vast amounts of human-written code. 
One has to assume that a lot of programming knowledge and practice that code evolution could exploit, is embodied in this treasure trove of correct examples. 
 
Another reason is to honor diverse approaches and recognize that any approach, in its early form, is going to be imperfect but may still be worthy of pushing its limits and improving it. 
No one knows precisely how LLM technology will advance and it is expedient that we  become familiar with its advantages and limitations now -- to either drive those advances or be set up for them when they arise.

Yet another reason is that working with LLMs stretches us intellectually and research should dig into their provocative novelty. 
LLMs offer a new computational paradigm, one working around pattern memory and matching. They do not offer a procedural abstraction. 
We, the GP community, are interested in the intelligence of Nature. 
We attend to nature's intelligent artifacts, including humanity and also social, collective, cooperative, competitive aspects of natural systems and organisms. 
How do the mechanisms of a LLM  relate to memory mechanisms within Natural systems?
Our community is also captivated by coding. 
Evolving code could lead to understanding the correspondence between an LLM's capabilities and Nature's mechanisms better. 
Could pattern completion competence be  effectively similar to highly environmentally-sensitive, self-adapted variation operators in the natural world?
Might it be a ``code-evolving'' scholar who discovers parallels between biological evolutionary mechanisms and an LLM? Might \GPLLM variants uncover insights into LLM capabilities that lead to advances in LLM design or usage, or GP approaches? Our community may also encounter answers as to how and why  LLMs retrieve code better than language.

We have stated the \GPLLM investigations skirt scientific method boundaries, but argued that there are still strong reasons to conduct them. 
In this case, how should investigations proceed?

\subsection{Conducting \GPLLM Investigations}\label{sec:discussion:conduct}
To date there are only 13 reports of \GPLLM of which only two are peer-reviewed. 
GP standards and norms started from a primitive state, i.e. they didn't exist on Day~1 and still evolve. 
Therefore, we could assume that committees and publication forums will provide the necessary encouragement for nascent work even with nascent standards. Here we suggest some possible initial standards:

\begin{description}
\item [Reporting]:
  \begin{itemize}
  \item  report the preparatory steps clearly.
  \item report time and cost of prompting during a run.
  \item report any biases beyond pre-training. 
  \item probe prompt sensitivity. If possible, also probe different LLMs. 
  \item maintain independent leaderboards on a benchmark for each of the GP and \GPLLM  approaches.
  \item try to pin down and report the exact model version along with its pre-training costs, its training data and its fine-tuning. 
  \end{itemize}

\item [Methods]:
  \begin{itemize}
  \item check the problem and solution are not in the data set!
  \item compare an LLM-based approach against other LLM-based  approaches when using a community benchmark. Consider whether (or not) it makes sense to compare with GP.
  \item make well-aligned comparisons (apples to apples, not apples to oranges). 
    GP  costs are incurred on different bases from \GPLLM. 
    Fitness evaluations dominate running cost so comparison among GP variants can be on the basis of the same number of fitness evaluations. But \GPLLM runs rely on a pre-trained model. The community needs a way of reconciling this cost when comparing. And,   the costs related to prompt response time and token cost have no GP equivalent. The  asymmetry remains to be addressed.
  \item address how much human intelligence has gone into solving the GP problem ahead of the \GPLLM run.  How would this differ in the case of GP, has it changed? Is domain information (not evolutionary information) hidden in a prompt?
  \end{itemize}

\item [Integrity]:
  \begin{itemize}
  \item be responsible with environmental cost. The budget devoted to investigation has the hidden expense of training an LLM. Multiple investigations with an open LLM could amortize its pre-training expense. 
  \item use the LLM ethically and keep usage aligned with human values.
  \end{itemize}
\end{description}

In the next section we propose some avenues for \GPLLM research.

\subsection*{Research Questions for \GPLLM}\label{sec:discussion:OpenRQs}

While it is infeasible to list all interesting \GPLLM avenues for future research, we offer:

\begin{description}
\item [Applications]:
\begin{itemize}
\item How can \GPLLM integrate software engineering domain knowledge? 
\item How can \GPLLM solve prompt composition or other LLM development and use challenges?
\item How can an EA using an LLM, but not necessarily evolving code, solve with different of units of evolution, e.g. strings, images, multi-modal candidates?
\end{itemize}
\item [Algorithm Variants]:
\begin{itemize}
\item How can we probe \GPLLM to understand the limits of its literal coding competence and more pragmatic coding competences?
\item How can a \GPLLM algorithm integrate design explorations related to cooperation, modularity, reuse, or competition?
\item How can a \GPLLM algorithm model biology differently from GP?
\item How can a \GPLLM intrinsically, or with guidance, support open-ended evolution?
\item What new variants hybridizing GP, \GPLLM and/or another search heuristic are possible and in what respects are they advantageous? 
\item Is there an elegant multi-objective optimization and many-objective optimization approach with \GPLLM? 
\end{itemize}
\item [Analysis Avenues]: 
\begin{itemize}
\item How well does \GPLLM scale with population size and problem complexity?
\item What is a search space in \GPLLM and how can it be characterized with respect to problem difficulty?
\item Does an LLM-based approach intrinsically address novelty or quality-diversity? To what extent, if so?
\item What is the most accurate computational complexity of \GPLLM?
\end{itemize}
\end{description}

%% file: conclusions.tex
\section{Conclusions \& Future Work}
\label{sec:conclusions--future}

This paper probes the novelty around using LLMs to evolve code. It provides clarity:
emerging from Algorithm~\ref{alg:gp_llm} is a sharp description of \GPLLM: it is an evolutionary algorithm that solves code synthesis. 
Like GP, it uses a set of evolutionary operators. 
However, its operators for initialization, selection, and variation can interface with an LLM via prompts that are returned from prompt-functions that are part of the hyper-parameters to the algorithm.
Like GP, its unit of evolution is code, but unlike GP's use of a structure that allows both execution and variation, it represents code with text that is a code snippet.
Emerging from an implementation and demonstration of execution, is the message that using an LLM to implement evolutionary operators incurs new costs: in a run, the time to interact with the LLM for each prompt, and the cost of prompting are significant. 
As well, a hidden, but significant cost is the pre-training of the LLM.

For practitioners, the paper provides a tutorial-level implemented variant of \GPLLM. 
It shows  the reader hands-on prompt function signatures, examples of prompt-functions and prompts and 
the module that interfaces with an LLM.
It explicates design decisions,  new hyper-parameters and new preparatory steps.
Finally, it contributes a discussion that up front itemizes the different risks and  uncertainties arising when using an LLM to evolve code. 
It then argues nonetheless for pursuing \GPLLM, primarily to not cut off potentially new insights.
It offers suggestions on how to conduct and report \GPLLM investigations and, to end, it offers avenues of potential investigation.

%% file: appendix.tex
\appendix

\section{Notation \& Definitions}\label{appdx-notation}

For notation see Table~\ref{tab:notation}.

\begin{table}
  \centering
  \caption{Notation and description. Note for readability we assume the context
    implies the dimensionality of some parameters. Note some symbols are overloaded and the context clarifies the use.}
  \label{tab:notation}
  \begin{tabular}{l|p{11cm}}
    \textbf{Symbol} & \textbf{Description}\\
    \hline
    \multicolumn{2}{c}{\textit{LLM}} \\
      \hline
      $A$ & Actor \\
    $\mathbf{x}$ & LLM Input. $\mathbf{x} \in \mathcal{T}$ \\ 
    $\mathbf{y}$ & LLM output. $\mathbf{y} \in \mathcal{T}$ \\ 
    $\Theta_{LLM}$ & LLM parameter and weights, $\Theta_{LLM} \in \mathbb{R}$ \\
      $f$ & A Large Language Model is parameterized model that
probabilistically outputs a sequence of tokens, $f: \mathcal{T} \times
\mathbb{R} \rightarrow \mathcal{T}, \mathbf{y} = f(\mathbf{x}|
\theta)$. \\ 

$g$ & A prompt is a function that outputs a sequence of tokens,
$g: \mathcal{T} \rightarrow \mathcal{T}, \mathbf{x}' =
g(\mathbf{x})$.\\
    \hline
    \multicolumn{2}{c}{\textit{GP}} \\
    \hline
    $g$ & generations~(iterations), $g \in \mathbb{N}$ \\
    $n$ & population size~(number of point samples), $n \in \mathbb{N}$\\
    $P^*$ & best solution, $P \in \mathcal{X}$\\
    $i$ & initialization function, $i: \mathbb{Z}^{\ge 0} \rightarrow \mathcal{X}, P = i(n)$\\
    $s$ & A selection function $s: \mathcal{X} \rightarrow \mathcal{X}, P' = s(P), P' \subseteq P$\\
    $v$ & A variation function $v: \mathcal{X} \rightarrow \mathcal{X}, P' = v(P)$\\
    $\rho_*$ & Prompt for $*$, $\rho \in \mathcal{T}$\\
    $c$ & A function for formatting an LLM response, $c: \mathcal{T}^m \times \mathcal{T} \rightarrow \mathcal{T}, r' = c(r,\cdot)$  \\
\hline
    $\mathbf{x}$ & A GP solution in the form of an expression tree, $\mathbf{x} \in \mathcal{G}, \mathcal{G} = (\mathcal{N},
    \mathcal{E})$ \\
    $n$ & A node in a GP tree. A node has a symbol $s$ and an
outdegree (arity $a$) $n = (s, a), s \in \mathcal{S}, a \in
\mathbb{Z}^{\geq}$\\
$s$ & A node symbol, $s \in \mathcal{S}$\\
$a$ & A node arity, $a \in
\mathbb{Z}^{\geq}$\\
    $|\mathbf{x}|$ & Number of nodes~(tree size) of a GP solution, $\mathbf{x} \in \mathbb{Z}^{\ge 1}$ \\
    $E$ & Executable environment, $E \in \mathcal{E}$\\
    $e$ & A function that evaluates a solution $e: \mathcal{P} \times \mathcal{E} \rightarrow \mathcal{Y}, \mathbf{y} = e(\rho, E)$ \\
$\phi$ & A measuring function of solution fitness~(quality) $\phi: \mathcal{Y} \times \mathcal{E} \rightarrow \mathbb{R}, \mathbf{f} = \phi(\rho, E)$. \\
   $D$ & Data, unlabeled or labeled\\
   $r$ & Response from LLM $f$ \\
   $\mathbf{a}$ & Prompt function arguments\\
  \end{tabular}
\end{table}

\section{LLM Operators}\label{sec:appdx-operators}

\begin{description}
\item[] \boldmath{$i_{LLM}$}, \textbf{Initialization:}\\ 
\textbf{Formulates} random candidate solutions using a prompt incorporating the primitives and instructions for formulating the solutions. \\
\textbf{Interfaces} to the LLM to execute the prompt and collect the response. \\
\textbf{Checks} response and reformulates if incorrect.


\item[]$e_{LLM}$,  \textbf{Execution} \\ 
\textbf{Formulates} prompt with instructions to execute a solution in an execution context, i.e. with data inputs.\\
\textbf{Interfaces} to the LLM to execute the prompt (and evaluate solution within it) and collect the response.\\
\textbf{Checks} response, returning an empty string if format is incorrect.\\
Note, LLM are notorious for not being able to compute mathematically so for problems requiring this, this operator is not used.

\item[] $ \phi_{\LLM}$ \textbf{Fitness measure}\\ 
\textbf{Formulates} prompt with instructions to use a fitness measure to assess the quality of the response/output from a prior evaluation of a solution.\\
\textbf{Interfaces} to the LLM to execute the prompt (and execute the measure on the prior response) and collect the response.\\
\textbf{Checks} response. An incorrect evaluation returns a default  value. \\
Note, LLM are notorious for not being able to compute mathematically so for problems requiring this, this operator is not used.

\item[] $s_{\LLM}$, \textbf{Selection}\\ 
\textbf{Formulates} prompt with instructions to select one solution in prompt over another. The prompt contains the solution and fitness for a list of
  individuals. The instructions are to select $k$ individuals from the list of
  individuals $\mathbf{p}$. Formally, $|\mathbf{p'}| = k, \mathbf{p'}
  \in \mathbf{p}$. \\
\textbf{Interfaces} to the LLM to execute the prompt and collect the response.
\textbf{Checks} response. An incorrect response returns individuals randomly  selected with replacement.\\ 
Note, the prompt size can be an error issue for  this formulation. This is due to the size of an individual and the  number of individuals (counted as tokens $mathcal{T}$.

\item[]\textbf{$V_{xoLLM}$}\textbf{, Crossover:}\\ 
\textbf{Formulates} prompt with instructions to recombine two parent solutions and instructions for combining them.\\
\textbf{Interfaces} to the LLM to execute the prompt and collect the response.\\
\textbf{Checks} response by extracting two child solutions.\\
In our implementation the prompt instructs the LLM to generate two new solutions that combine elements of the two solutions but it does not check whether all elements appear across the two new solutions. 

\item[] $r_{LLM}$, \textbf{Replacement}\\  
\textbf{Formulates} prompt with instructions to rank a set of given solutions and select the  best. \\
\textbf{Interfaces} to the LLM to execute the prompt and collect the response.\\
\textbf{Checks} response.

\item[] $b_{LLM}$, \textbf{Ranking solutions} \\ 
\textbf{Formulates} prompt with instructions to rank a set of given solutions and select the  best. \\
\textbf{Interfaces} to the LLM to execute the prompt and collect the response.\\
\textbf{Checks} response.

\end{description}

\section{Prompts}
\label{sec:prompts}

\paragraph{Prompt implementations}

Prompts are in Python syntax. \texttt{\{,\}} indicates variable substitution when formatting a string. \texttt{"""} indicates a string.

The primitives are the allowed GP symbols. Few shot samples are the taken from the current population.

\subsubsection{\GPLLM  Few shot symbolic regression prompts using Primitives}

{\scriptsize
\begin{verbatim}
SORT_POPULATION_PROMPT_FEW_SHOT = """ An example of an order is in the
following list: {samples}

Order the elements of the following list: {individuals}

Provide no additional text in response. Format output in JSON as
{{"individuals": ["<element>"]}} """

SELECTION_PROMPT_FEW_SHOT = """ {n_samples} examples of high quality
elements are: {samples}

Select {population_size} elements of high quality from the following
list: {individuals}

Provide no additional text in response. Format output in JSON as
{{"individuals": ["<element>"]}} """

REPLACEMENT_PROMPT_FEW_SHOT = SELECTION_PROMPT_FEW_SHOT

REPHRASE_MUTATION_PROMPT_FEW_SHOT = """ {n_samples} examples of
mathematical expressions are: {samples}

Rephrase the mathematical expression {expression} into a new
mathematical expression. Use the listed symbols {constraints}.

Provide no additional text in response. Format output in JSON as
{{"new_expression": "<new expression>"}} """

CROSSOVER_PROMPT_FEW_SHOT = """ {n_samples} examples of mathematical
expressions are: {samples}

Recombine the mathematical expressions {expression} and create
{n_children} new expressions from the terms. Use only the the existing
expression when creating the new expressions.

Provide no additional text in response. Format output in JSON as
{{"expressions": ["<expression>"]}} """

EVALUATION_PROMPT_FEW_SHOT = """ {n_samples} examples outputs from
mathematical expressions are: {samples}

Provide the output from the evaluation of the mathematical expression
{expression} on the following list of variables: {exemplars}

Provide no additional text in response. Format output in JSON as
{{"outputs": ["<output>"]}} """

FITNESS_MEASURE_PROMPT_FEW_SHOT = """ {n_samples} examples numerical
quality scores for mathematical expressions are: {samples}

Provide a numerical quality score based on the list of outputs from
expression {expression}: {outputs} When comparing it to the following
list of targets: {targets}

Provide no additional text in response. Format output in JSON as
{{"fitness": "<quality>"}} """
\end{verbatim}
}

\subsubsection{LLM one  symbolic regression prompt}

{\scriptsize
\begin{verbatim}
PROMPT = """ Generate a mathematical expression for the following
variables.  {exemplars}

Use the operators: {constraints}.

Provide no additional text in response. Format output in JSON as
{{"expression": "<expression>"}} """
\end{verbatim}
}